\documentclass[10pt]{article} 

\usepackage[accepted]{tmlr}

\usepackage{amsmath,amsfonts,bm}

\def\eqref#1{equation~\ref{#1}}

\def\1{\bm{1}}

\DeclareMathAlphabet{\mathsfit}{\encodingdefault}{\sfdefault}{m}{sl}
\SetMathAlphabet{\mathsfit}{bold}{\encodingdefault}{\sfdefault}{bx}{n}

\usepackage{hyperref}
\usepackage{url}

\usepackage{graphicx}
\usepackage{booktabs}
\usepackage{multirow}
\usepackage{amsmath}
\usepackage{amssymb}
\usepackage{enumitem}
\usepackage{caption}
\usepackage{wrapfig}
\usepackage[table,dvipsnames]{xcolor}
\definecolor{hlblue}{RGB}{168,222,255}
\definecolor{hlyellow}{RGB}{255,245,168}
\definecolor{hlred}{RGB}{255,185,168}
\usepackage{listings}
\usepackage[most]{tcolorbox}
\tcbuselibrary{breakable}
\tcbuselibrary{skins}
\tcbset{enhanced, attach boxed title to top left={yshift=-3mm,yshifttext=-1mm, xshift=3mm}, boxed title style={size=small}, colback=white!0}
\usepackage{soul}

\usepackage{subcaption}   

\newtcblisting{promptbox}[2][]{
  enhanced,
  breakable,
  listing only,
  listing options={
    basicstyle=\ttfamily\small,
    breaklines=true,
    columns=fullflexible,
    keepspaces=true,
    showstringspaces=false
  },
  colback=gray!4,
  colframe=gray!55,
  coltitle=gray!10,
  fonttitle=\bfseries,
  title={#2},
  boxrule=0.6pt,
  arc=2pt,
  left=6pt,
  right=6pt,
  top=6pt,
  bottom=6pt,
  #1
}

\usepackage{minitoc}

\title{Thinking Is Not Telling: Information Disclosure in User-Service LLM Agents}

\author{\name Jiatong Li \email jli2947@cs.wisc.edu \\
      \addr University of Wisconsin-Madison
      \AND
      \name Changdae Oh \email changdae@cs.wisc.edu \\
      \addr University of Wisconsin-Madison
      \AND
      \name Hyeong Kyu Choi \email froilanchoi@cs.wisc.edu \\
      \addr University of Wisconsin-Madison
      \AND
      \name Jindong Wang \email jdw@wm.edu \\
      \addr William \& Mary
      \AND
      \name Sharon Li \email sharonli@cs.wisc.edu \\
      \addr University of Wisconsin-Madison
    }

\def\month{10}  
\def\year{2026} 
\def\openreview{\url{https://openreview.net/forum?id=xngXTG9Jyj}} 

\begin{document}

\maketitle

\begin{abstract}
User-engaged LLM agents increasingly operate in service scenarios where task success depends on coordination between the agent, the user, and a stateful environment. In such interactions, the agent often has access to rich information that the user does not observe, such as task policies, tool results, and environment states. This makes agent-user communication a central component of task completion. In this work, we study a communication-level failure mode of user-service agents: 
internal reasoning is counterproductive for the interactive information disclosure.
We introduce \textit{information disclosure} as a measurable communication mechanism, covering state corrections, available options, constraints, consequences, and proactively retrieved environment information. Motivated by recent advances in LLM reasoning, we evaluate and analyse information disclosure under a controlled setting of agents equipped with a ``think'' tool, which allows agents to reason internally before responding or acting. Across five models and three user-service scenarios, we first show that enforced thinking has unstable utility. We then analyse response scaling under different think efforts and find that more thinking does not reliably translate into richer information disclosure. Using evidence-grounded probing and turn-level counterfactual intervention, we show that injecting golden decision-relevant information significantly intervenes agents to correct action trajectories, indicating that missing disclosure can causally affect the agent's immediate downstream environment-edit behaviour. We further find that failed trajectories most often lack state corrections and constraints, and that a lightweight disclosure-aware prompt effectively mitigates degradation from enforced thinking. These results suggest that the key challenge is not simply making agents think more, but ensuring that reasoning and environment information are externalised as timely, user-facing information disclosure. Code is available at \url{https://github.com/deeplearning-wisc/Agent-InfoDis}.

\end{abstract}

\section{Introduction}

\par Large language model (LLM) agents are increasingly applied to user-service scenarios where task completion requires coordination among an agent, a user, and a stateful environment~\citep{yao2023react,gao2025surveyselfevolvingagentspath}. Unlike conventional reasoning benchmarks in which the full problem specification is given upfront, user-engaged tasks often unfold as collaborative decision-making under asymmetric and partial information~\citep{hemmer2025complementarity,luo2025clarifymtbench, lu2025toolsandbox, wang2024mint, qian2025userbench}. The agent may have access to rich hard-to-read information including task policies, tool interfaces, and environment states, while the user holds complicated intent: goals, preferences, or constraints. As a result, the agent must establish sufficient common ground with the user before consequential decisions are made.

\par In previous studies, a common practice in enhancing LLM agents is to equip them with higher reasoning efforts~\citep{yao2024tau,anthropic2025think}, which allows agents to handle complicated intent understanding, task planning and logical reasoning~\citep{muennighoff2025s1,wei2022cot}. However, the interaction-aware characteristic of user-engaged scenarios makes the contribution of internal reasoning doubtful. In our large-scale validation across five models, three user-service scenarios, and two think modes, enforcing internal thinking does not reliably improve agent performance (see Table~\ref{tab:performance-comparison}). This raises a mechanism-level research question: 

\textbf{How productive is thinking for incentivising agents' user-service behaviour?}

\par We argue that the missing piece lies in the distinction between \textit{external communication} and \textit{internal reasoning}. In user-engaged tasks, the agent's success is directly decided by whether it has sufficient communication with the user and whether it applies correct environment-aware actions during the communication. The agent's internal reasoning is useful only insofar as it is converted into appropriate user-facing responses and environment actions. When the next state transition depends on the user's decision -- confirmation, selection, or preference~\citep{williams2007partially}, the user must be informed of the relevant information: state, options, constraints, and consequences~\citep{clark1991brennan}. Although an internal think step may help the agent organise its own plan, it does not guarantee that decision-relevant information is either identified by the agent or disclosed to the user at the right time. This view is consistent with theories of shared collaborative plans~\citep{grosz1996collaborative}, POMDP-based models of dialogue, and cooperative decision-making under uncertainty~\citep{hadfield2016cooperative,lin2024decision}. As a result, a natural communication--reasoning mismatch exists: the user still makes decisions under incomplete information while the agent may privately process useful information.

\par To study this mismatch, we introduce \textbf{information disclosure} as a measurable communication mechanism in user-engaged LLM agents. We define information disclosure as the agent's proactive communication of decision-relevant information that the user likely does not possess but needs for the next decision. This includes state corrections, available options, task constraints, decision consequences, and proactively retrieved environment information. This perspective complements prior work on information clarification~\citep{luo2025clarifymtbench, zhang2025clarify, suri2026structured}, where the agent elicits missing information from the user. In contrast, information disclosure studies the reverse direction of information flow: what the agent should reveal to help the user make an informed decision.
\begin{figure}[t]
    \centering
    \includegraphics[width=\linewidth]{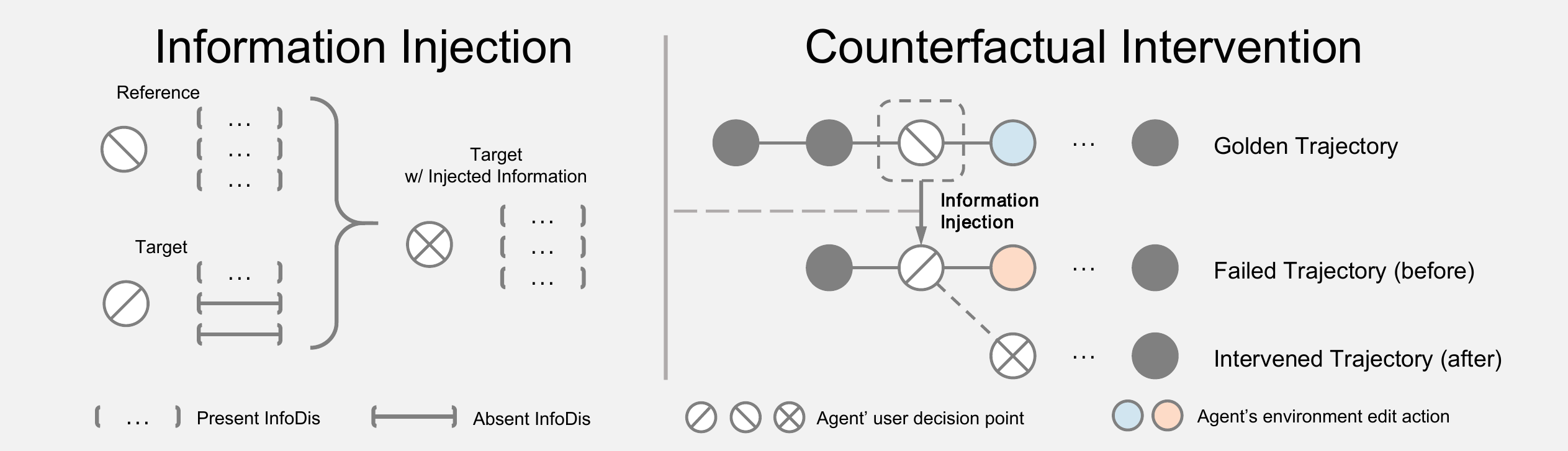}
    \caption{An overview of the counterfactual intervention design. To handle intervention in multi-turn dialogue processes, the method identifies the \textit{last} agent's user decision point before the \textit{first} environment edit action. Next, the method injects the InfoDis evidence annotated from evidence-grounded probing.}
    \label{fig:counterfactual_overview}
\end{figure}

\par We conduct a multi-stage empirical study to examine how information disclosure and downstream task dynamics are affected by interal thinking efforts. Taking the ``think'' tool as a controlled and practically important instantiation, we first evaluate across multiple models and scenarios, showing that enforced thinking does not consistently improve user-service task success. Second, we analyse different levels of think effort and find that more internal thinking does not reliably translate to user-facing communication. Third, a motivating case study shows that an enforced-think agent can omit a critical task constraint before the user's decision, misleading the user toward an invalid action. We then formalise this phenomenon through evidence-grounded probing of information disclosure and perform turn-level counterfactual interventions on failed trajectories (see Figure~\ref{fig:counterfactual_overview}). Compared with verbosity and think-content controls, injecting golden decision-relevant information produces a more significant improvement in First Edit Correction Rate (FECR), indicating that \textit{missing disclosure can directly affect the agent's immediate downstream environment-edit behavior}. Finally, we show that a simple InfoDis-aware prompting strategy mitigates the degradation of enforced thinking and improves overall performance across models and domains. Overall, this study demonstrates that information disclosure is an interpretable and controllable factor of user-engaged LLM agents, providing insights for future development of relevant fields. 
\par In summary, this study makes three contributions:

\begin{itemize}[leftmargin=*]
    \item We formulate missing decision-relevant information disclosure as a general communication bottleneck in user-service LLM agents and show that augmenting internal reasoning alone does not ensure that users receive the information required for informed decisions.
    \item We introduce a five-category taxonomy and an evidence-grounded framework for probing information disclosure in agent-user dialogues. Through turn-level counterfactual intervention, we demonstrate that missing disclosure is an interpretable and actionable factor in downstream agent failures.
    \item We evaluate disclosure-aware prompting as a lightweight, training-free intervention for both thinking and non-thinking agents, showing that explicitly optimizing user-facing disclosure can improve overall performance and, in particular, mitigate degradation under enforced thinking.
\end{itemize}

\section{Related Work}

\paragraph{Agent-user communication.}
User-engaged agents have become an important class of LLM agents in both industrial and academic settings~\citep{clarke2022one,deshpande2025multichallenge}. These agents communicate with users through natural language~\citep{qian2024tell} while interacting with external environments through formatted actions or tool calls~\citep{yao2023react,openai2025function}. Accordingly, recent benchmarks have moved beyond static multi-turn dialogue evaluation~\citep{bai2024mtbench101} toward interactive evaluation with simulated users and verifiable task outcomes, such as database-state matching in $\tau$-Bench~\citep{yao2024tau}, tool-use feedback in MINT~\citep{wang2024mint}, and milestone tracking in ToolSandbox~\citep{lu2025toolsandbox}. A key challenge in user-agent communication is asymmetric information. Existing studies mainly address this issue through \textit{information clarification}, where the agent elicits missing or ambiguous information from the user. Prior works cover implicit user intention understanding~\citep{qian2024tell}, deciding when to ask clarification questions~\citep{zhang2025clarify}, clarification-oriented dialogue evaluation~\citep{luo2025clarifymtbench}, and training agents to reason, speak, and act interactively~\citep{acikgoz2025speakrl}. In contrast, our work studies the reverse direction of information flow: \textit{information disclosure}, where the agent proactively communicates task-relevant information to help the user make informed decisions. This perspective complements clarification-based interaction and highlights a different failure mode: the agent may have access to information that the user needs, but fail to disclose it before the user's next decision.

\paragraph{Reasoning of LLM agents.}
Reasoning has been widely studied as a core capability of LLMs and LLM agents. Chain-of-thought prompting improves mathematical and symbolic reasoning by eliciting intermediate reasoning traces~\citep{wei2022cot,kojima2022large}, and later methods extend this idea through structured search, planning, and reflection~\citep{yao2023tree,shinn2023reflexion,hao2023reasoning}. In agent settings, ReAct interleaves reasoning traces with environment actions~\citep{yao2023react}, and subsequent agent frameworks explore reflection, memory, and planning for long-horizon tool use~\citep{shinn2023reflexion,wang2024survey}. However, explicit reasoning does not always improve user-engaged tool-calling agents. Previous works in reasoning have discovered the overthinking problem~\citep{ghosal2025does}. Recent studies find that chain-of-thought or reasoning traces can have marginal or inconsistent utility in interactive tool-use settings~\citep{yao2024tau,sprague2025cot,zhang2026does}. Meanwhile, the private ``think'' tool has been introduced as a model-agnostic way for tool-calling agents to write intermediate reasoning into their own context without exposing it to users or modifying the environment~\citep{anthropic2025think}. Existing explanations for its unstable utility often focus on insufficient training~\citep{wang2024llms}, hallucinated thoughts~\citep{yin2026reasoning, baker2025detecting}, or mismatch between reasoning and tool-use objectives~\citep{sprague2025cot}. Our work instead examines a communication-level mechanism: in user-engaged tasks, internal reasoning may fail to help when decision-relevant information is not externalized to the user before the next decision point.

\section{An Empirical Study of Think on Response Scaling}

\subsection{Preliminary Evaluation of the ``Think'' Tool}
We consider a user-engaged tool-calling agent that interacts with a user through natural language and with an external environment through tool calls. Let $F=\{f_1,\ldots,f_C\}$ denote the original tool set, and let $f_{\text{think}}$ denote a internal reasoning tool. Following prior practice~\citep{anthropic2025think, yao2024tau}, $f_{\text{think}}$ does not query the environment or modify its state; instead, it appends a free-form reasoning trace to the agent's internal context. This trace is visible to the agent in subsequent steps but invisible to the user. We compare agents equipped with $F$ against agents equipped with $F\cup\{f_{\text{think}}\}$. As a preliminary validation, we evaluate whether enforcing the ``think'' tool before each agent step improves task performance. 
\par Our evaluation covers diverse model families and generations, including GPT-4o~\citep{openai2024gpt4o}, GPT-4.1~\citep{openai2025gpt41}, GPT-5~\citep{openai2025gpt5}, DeepSeek-V3.1~\citep{deepseek2025deepseekv31}, and gpt-oss-120b~\citep{openai2025gptoss120b}. We use three representative user-service scenarios: Retail and Airline from $\tau$-Bench~\citep{yao2024tau}, and the multi-turn phone scenario from ToolSandbox, denoted as TS-Phone~\citep{lu2025toolsandbox}. The original ToolSandbox uses a process-aware milestone similarity in $[0,1]$ as the evaluation metric. For our comparative evaluation, we label similarity > 0.9 as ``passed'' to get a binary outcome score. Each setting is evaluated with three random seeds, and we report pass\^{}1 and pass\^{}2, where pass\^{}k measures the probability that all $k$ independent trials of a task succeed. Table~\ref{tab:performance-comparison} shows that enforced thinking does not consistently improve user-engaged tool-calling agents. While it helps in a few settings, such as gpt-oss-120b on Retail and GPT-4.1 on Airline, it also substantially degrades performance in others, such as GPT-4.1 on Retail, where pass\^{}2 drops from $63.48\%$ to $52.46\%$, and GPT-4o on Airline, where pass\^{}2 drops from $30.00\%$ to $19.33\%$. These results suggest that, unlike reasoning-centric benchmarks, where extended reasoning often improves performance, it does not necessarily improve user-service agents. 

\begin{table}[t]
\caption{Performance comparison of user-engaged tool-calling agents under different think tool mode. Each setting is run for three trials with random seeds. The pass\^{}k quantifies the chance that all $k$ i.i.d. task trials are successful, averaged across tasks, representing the reliability and consistency of LLM agents. }~\label{tab:performance-comparison}
\begin{center}
\resizebox{0.9\linewidth}{!}{
\begin{tabular}{lr|rrrrrr}
\toprule
\multirow{2}{*}{\textbf{Model}} & \multirow{2}{*}{\textbf{Think Mode}} & \multicolumn{2}{c}{\textbf{Retail}} & \multicolumn{2}{c}{\textbf{Airline}} & \multicolumn{2}{c}{\textbf{TS-Phone}} \\
 &  & \textbf{Pass\^{}1} & \textbf{Pass\^{}2} & \textbf{Pass\^{}1} & \textbf{Pass\^{}2} & \textbf{Pass\^{}1} & \textbf{Pass\^{}2} \\
 \midrule
\multirow{2}{*}{GPT-4o} & w/o think & \textbf{64.06} & \textbf{53.33} & \textbf{43.33} & \textbf{30.00} & 39.29 & 30.21 \\
 & Enforced & 59.71 & 45.51 & 35.33 & 19.33 & \textbf{41.52} & \textbf{31.85} \\  \midrule
\multirow{2}{*}{GPT-4.1} & w/o think & \textbf{73.33} & \textbf{63.48} & 44.00 & 31.33 & \textbf{46.73} & \textbf{37.65} \\
 & Enforced & 65.51 & 52.46 & \textbf{52.00} & \textbf{40.67} & 45.24 & 35.57 \\  \midrule
\multirow{2}{*}{GPT-5} & w/o think & \textbf{74.78} & \textbf{64.78} & \textbf{50.67} & \textbf{42.67} & 24.85 & 15.77 \\
 & Enforced & 69.57 & 56.23 & 42.00 & 32.00 & \textbf{28.72} & \textbf{19.64} \\  \midrule
\multirow{2}{*}{DeepSeek-V3.1} & w/o think & \textbf{71.01} & \textbf{59.71} & \textbf{50.00} & 34.67 & 41.52 & \textbf{30.80} \\
 & Enforced & 63.48 & 46.09 & 49.33 & \textbf{35.33} & \textbf{41.82} & 29.91 \\  \midrule
\multirow{2}{*}{gpt-oss-120b} & w/o think & 63.77 & 51.59 & \textbf{45.33} & \textbf{31.33} & \textbf{38.10} & \textbf{27.68} \\
 & Enforced & \textbf{66.67} & \textbf{55.65} & 40.67 & 27.33 & 37.05 & 25.15 \\ \bottomrule
\end{tabular}}
\end{center}
\end{table}

\begin{figure}[t]
    \centering
    \includegraphics[width=\linewidth]{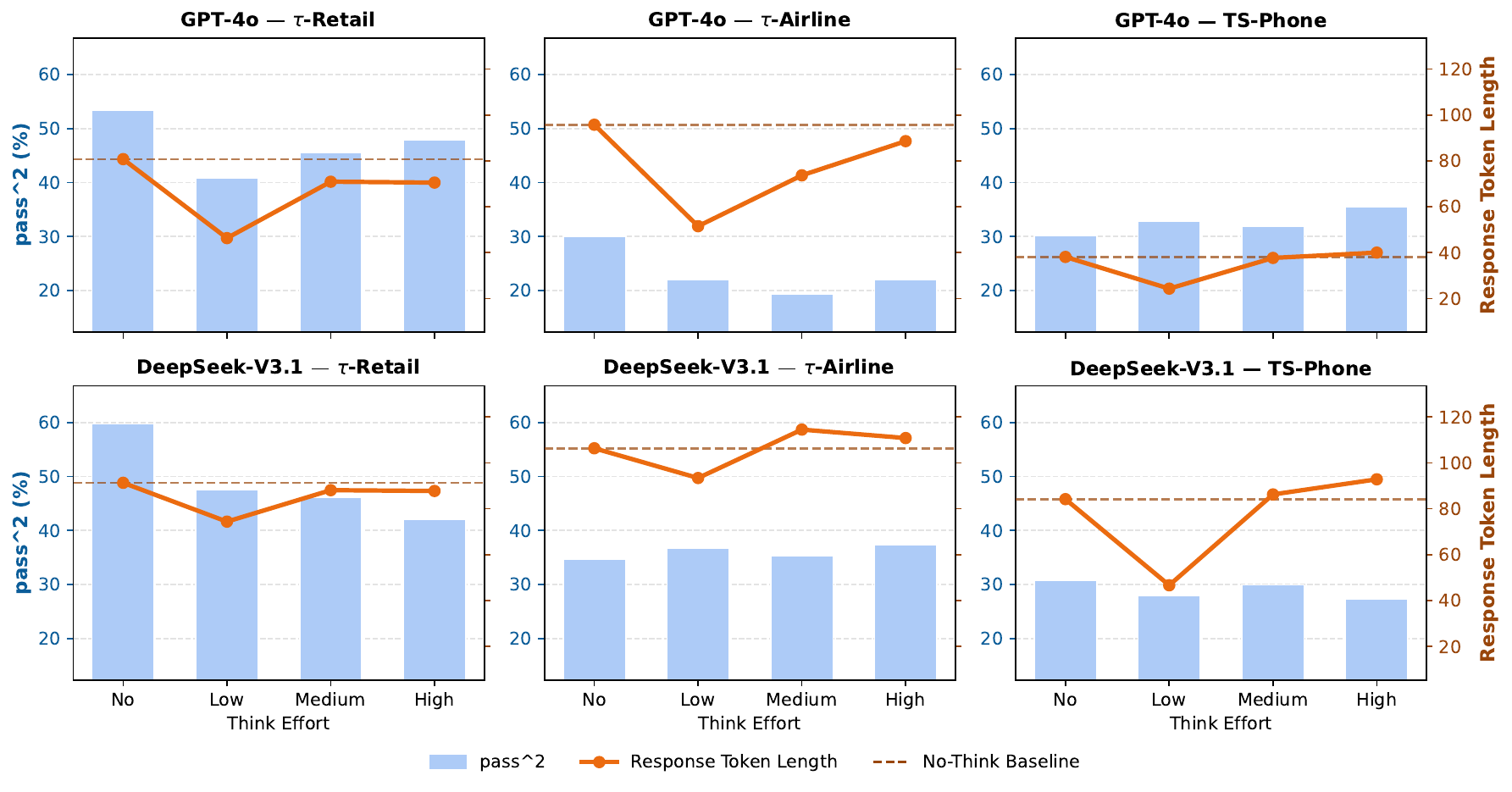}
    \caption{Response token length and pass\^{}2 with respect to think effort.}
    \label{fig:response-scaling}
\end{figure}

\subsection{Performance and Response Scaling Comparison}
\par To examine how the think tool affects user-service agents, we first study \textit{response scaling} under different think efforts. In user-engaged settings, think content is private to the agent, while user-visible responses are the only communication channel available to the user. Thus, response length serves as a coarse surface proxy for how much information is communicated. Following prior work, we compare four settings: \textit{no think}, \textit{low think} (System 1), \textit{medium think} (default), and \textit{high think} (System 2)~\citep{kahneman2011thinking,li202512surveyreasoning}. We append System 1/2 instructions to the agent system prompt to control low/high thinking efforts, with details in Appendix~\ref{app:prompt}. As shown in Figure~\ref{fig:response-scaling}, increasing internal think effort does not reliably improve performance or scale user-facing responses. In several settings, response length even decreases despite enforced thinking. This observation suggests that tool-based internal reasoning does not automatically translate into richer user-facing communication. We next examine this effect at the semantic level through a case study of agent outputs.

\begin{figure}[t]
\begin{minipage}{0.45\linewidth}
    \begin{tcolorbox}[title={Agent w/o think},left=1.2mm, right=1.2mm,top=1.4mm, bottom=0.2mm]\footnotesize

\begin{lstlisting}
Hi, I'd like to return and cancel some items.
\end{lstlisting}

...

\vspace{0.5em}
{\color{gray!40}\hrule height 2pt}
\vspace{0.5em}

\textbf{Agent}
\begin{lstlisting}
To confirm, you want to cancel only the Garden Hose (black, 50ft, vinyl, Item ID: 5206946487) from your pending order #W2575533, with the reason "no longer needed." 
(*{\sethlcolor{hlblue}\hl{Please note: I can only cancel the entire pending order, not individual items. If you proceed, the whole order \#W2575533 (including all items in it) will be cancelled and refunded to your PayPal account in 5–7 business days.}}*) 
Would you like to cancel the entire order #W2575533, or would you prefer to keep the other items? Please confirm how you want to proceed.
\end{lstlisting}

\vspace{0.2em}
{\color{gray!40}\hrule height 2pt}
\vspace{0.5em}

\textbf{User}
\begin{lstlisting}
I'd prefer to keep the other items, (*{\textbf{so let's skip cancelling.}*) Just return the other items. How much refund total?
\end{lstlisting}
\end{tcolorbox}
\end{minipage}
\hfill
\begin{minipage}{0.53\linewidth}
    \begin{tcolorbox}[title={Agent with enforced think},left=1.2mm, right=1.2mm,top=1.4mm, bottom=0.2mm]\footnotesize

\begin{lstlisting}
Hi, I need help returning a few items and cancelling an item.
\end{lstlisting}

...

\vspace{0.5em}
{\color{gray!40}\hrule height 2pt}
\vspace{0.5em}

\textbf{Agent}
{%
\lstset{
    basicstyle=\scriptsize,
    backgroundcolor=\color{gray!20},  
    frame=none,                       
    framexleftmargin=3pt,             
    framexrightmargin=3pt,            
    framextopmargin=3pt,
    framexbottommargin=3pt,
}
\begin{lstlisting}
(*\textbf{[Thought]}*) She wants to return items from delivered orders (#W3792453, #W7181492, #W5565470) and cancel a pending garden hose order (#W2575533). I need to confirm the details with her before proceeding, including the payment method for the return refund and the reason for cancellation.
\end{lstlisting}
}%
\begin{lstlisting}
Here are the details I found for your requests:
1. Return items from delivered orders: ...
2. Cancel a pending order: Garden Hose (#W2575533)
To proceed, please confirm: ...
- You want to cancel the pending garden hose order (#W2575533) because it is 'no longer needed' or 'ordered by mistake' 
\end{lstlisting}

\vspace{-0.2em}
{\color{gray!40}\hrule height 2pt}
\vspace{0.5em}

\textbf{User}
\begin{lstlisting}
Yes, that's correct. Please refund to my original PayPal. (*\textbf{The hose order should be cancelled because it was ordered by mistake.}*)
\end{lstlisting}
\end{tcolorbox}
\end{minipage}\hspace{2pt}

\caption{A case of agent-user interaction before the first environment edit action on a retail task (agent: GPT-4o; user: GPT-4.1). The {\sethlcolor{hlblue}\hl{blue}} highlights the information disclosure of task \textbf{constraint}, which intervenes the next user decision. The thought of agent with enforced think is invisible to the user.}\label{fig:case-study}

\end{figure}
\subsection{Case Study on Agent Output}

\par Figure~\ref{fig:case-study} shows a representative failure pattern before the first environment-edit action. Both agents face the same retail task that helps a user to return and cancel items. In the trajectory without the think tool, the agent explicitly discloses a key \textit{task constraint} that the system can only cancel the entire pending order rather than an individual item. This constraint changes the user's next decision: the user chooses to skip the cancellation and proceed only with the return. In contrast, the agent with enforced think internally records relevant task context in its private thought, but its user-visible response omits the constraint. The agent lists the options and asks the user to confirm the cancellation reason, without informing the user of the cancellation constraint. Therefore, the user makes the next decision under incomplete information and confirms the cancellation. We refer to this missing user-facing communication as \textbf{information disclosure} (InfoDis). In user-engaged tasks, InfoDis denotes the agent's proactive communication of decision-relevant information that the user likely does not possess but needs before making the next decision. Figure~\ref{fig:case-study} presents a \textit{constraint disclosure}. More broadly, information disclosure can include state correction, available options, constraints, consequences, and proactively retrieved environment information. In the next section, we formalize these dimensions and test their effects through turn-level counterfactual intervention.

\begin{table}[t]
\caption{Information disclosure dimensions.}\label{tab:infodis-rubrics}

\begin{center}
\resizebox{\linewidth}{!}{\begin{tabular}{lll}
\toprule
\textbf{Category} & \textbf{Definition} & \textbf{Example} \\
\midrule
State Correction (SC) & \begin{tabular}[c]{@{}l@{}}Information that corrects a factual misunderstanding or \\ false assumption the user has expressed.\end{tabular} & \begin{tabular}[c]{@{}l@{}}The User says "my cancelled order" but agent \\ clarifies "this order was delivered, not cancelled"\end{tabular} \\ \midrule
Option (OP) & \begin{tabular}[c]{@{}l@{}}The set of choices available to the user for their upcoming \\ decision — including relevant variants, alternatives, or paths.\end{tabular} & \begin{tabular}[c]{@{}l@{}}``Here are available options for replacement: \\ - ...\\ - ...''\end{tabular} \\ \midrule
Constraint (CN) & \begin{tabular}[c]{@{}l@{}}Rules, policies, eligibility conditions, or limits that affect \\ what the user can or cannot do — and that the user likely \\ does not already know.\end{tabular} & ``Modifications are only possible before shipment'' \\ \midrule
Consequence (CQ) & \begin{tabular}[c]{@{}l@{}}The concrete outcomes that will result from the user's decision, \\ including financial impact, timing, and irreversible effects.\end{tabular} & \begin{tabular}[c]{@{}l@{}}Refund amount, refund timeline, charge amount, \\ order status change\end{tabular} \\ \midrule
Proactive Retrieval (PR) & \begin{tabular}[c]{@{}l@{}}Information the agent proactively fetched from its tools\\ /database and disclosed even though the user did not possess \\ and did not directly ask for.\end{tabular} & \begin{tabular}[c]{@{}l@{}}Listing the user's orders with their contents, showing \\ both current and new address for comparison\end{tabular} \\
\bottomrule
\end{tabular}}
\end{center}
\end{table}

\section{Counterfactual Intervention Pipeline of Information Disclosure}
\label{sec:experiment}

\subsection{Evidence-Grounded Semantic Probing of Information Disclosure}
\label{sec:probing}

\par To study information disclosure in agent-user interaction, we first need to make it observable and measurable. This is challenging because information disclosure is expressed through natural language rather than through a structured action space. The same response may contain multiple types of task-relevant information, and its meaning depends on the preceding user request, retrieved environment states, task policies, and the upcoming user decision. Moreover, response length alone is insufficient: a short response may disclose a critical constraint, while a long response may add verbosity without changing what the user knows. We therefore avoid assigning a scalar ``informativeness'' score and instead use an evidence-grounded probing procedure, where an information category is marked as present only when an exact text span in the user-visible response supports the label. The bracketed information lists in the left part of Figure~\ref{fig:counterfactual_overview} illustrate this probing process.

\paragraph{Information Disclosure Rubrics.}
We operationalize information disclosure with five binary categories, summarized in Table~\ref{tab:infodis-rubrics}: \textit{State Correction (SC)}, \textit{Option (OP)}, \textit{Constraint (CN)}, \textit{Consequence (CQ)}, and \textit{Proactive Retrieval (PR)}. These categories capture common forms of decision-relevant information in user-service tasks: correcting the user's false assumptions, presenting feasible choices, disclosing rules or eligibility conditions, explaining downstream outcomes, and revealing environment information proactively retrieved by the agent. For each agent response, every category is annotated as either present or absent. We use binary labels rather than graded scores to reduce ambiguity and to make trajectory-level information disclosure comparable across models, domains, and intervention conditions.

\paragraph{Evidence-Grounded Annotation.}
Given an agent-user trajectory, we perform annotation at the agent-turn level. To avoid exposing annotators to experimental conditions, we mask internal reasoning traces and tool calls, and present only the system prompt, user utterances, and user-visible agent responses. For each agent response, the annotator labels the five InfoDis categories. A category can be marked as present only if the annotator identifies an exact substring in the response that directly supports the label; otherwise, it is marked as absent. We also annotate whether each agent turn is a \textit{user decision point}, defined as a turn after which the next user response can determine the subsequent task trajectory, such as confirming an action, selecting among options, providing preferences, or authorizing an environment edit. These annotations provide the basis for comparing successful reference trajectories with failed target trajectories and identifying missing disclosure evidence for counterfactual injection, as shown in Figure~\ref{fig:counterfactual_overview}.

\subsection{Counterfactual Information Injection}
\label{sec:intervention}

\par Figure~\ref{fig:counterfactual_overview} illustrates our turn-level counterfactual intervention pipeline. Given a failed target trajectory and a matched successful golden trajectory for the same task, we first compare their evidence-grounded InfoDis annotations. The goal is to identify decision-relevant information that is present in the golden response but absent from the failed response, and then inject this missing information into the failed trajectory before the agent makes its first task-critical environment edit. Formally, let $\tau$ be a failed target trajectory and $\tau^*$ be a matched successful golden trajectory. Both trajectories are annotated with InfoDis labels and evidence spans as described in Section~\ref{sec:probing}. For the target trajectory, we locate the first environment-edit action, such as order cancellation, refund creation, flight change, or reservation update. We then select the last user decision point before this edit as the intervention turn. This choice targets the moment where additional disclosure can still affect the user's next response and the agent's subsequent edit action. 

Next, the injected content is constructed by comparing the target response with the corresponding golden response at the selected decision point. InfoDis categories that are supported by evidence in the golden response but absent from the target response are treated as missing disclosure. We use the golden evidence spans as the source of injected information. To avoid unnatural concatenation, a rewriting model minimally revises the original target response so that the missing evidence is naturally included while preserving the response intent, style, and next question.

Finally, the rewritten response replaces the original target response, and the dialogue is rolled out from that point with the same agent and user simulator. We evaluate whether the intervention changes the first environment-edit action and whether the final trajectory outcome matches the golden trajectory. To separate the effect of decision-relevant disclosure from generic text perturbation, we compare \texttt{+Golden Info} against two controls: \texttt{+Think Content}, which injects content extracted from the agent's internal think trace, and \texttt{+Verbosity}, which increases response verbosity without adding the missing golden evidence.

\section{Experiments}
\par We aim to answer three research questions in this experiment:
\begin{itemize}[leftmargin=*]
    \item RQ1: What is the causal effect of injecting information by different methods?
    \item RQ2: Which type of information disclosure do failed trials really lack?
    \item RQ3: Can we boost agent performance by improving their information disclosure?
\end{itemize}

\subsection{Experiment Setup}
\paragraph{Environments and Models.}
Considering the engagement of user and task complexity, for agent scenarios, we conduct our study on the Retail and Airline on the $\tau$-Bench. For agent LLMs, we take two representative from different model families: GPT-4o and DeepSeek-V3.1. For user simulator, we take GPT-4.1 considering its general capability, following the evaluation in \citet{yao2024tau}. Regarding environment and decoding setting, the \texttt{temperature} is set as 0.2 for the agent and 1.0 for the user, respectively. The \texttt{max\_token} is set to 4096 for both LLMs. To mitigate randomness from the sampling of agent and user outputs, each agent task is run for three trials for any setting.

\begin{table}[t]
\caption{Summary of processed data for counterfactual intervention.}\label{tab:filter-data}
\begin{center}
\resizebox{0.6\linewidth}{!}{\begin{tabular}{llrrr}
\toprule
\textbf{Model} & \textbf{Scenario} & \textbf{Failed Trajs} & \textbf{Golden Trajs} & \textbf{Filtered Links} \\ \midrule
GPT-4o & Retail & 73 & 48 & 64 \\
GPT-4o & Airline & 48 & 25 & 44 \\
DeepSeek-V3.1 & Retail & 91 & 58 & 63 \\
DeepSeek-V3.1 & Airline & 48 & 27 & 43 \\ \midrule
\multicolumn{2}{l}{Total} & 260 & 158 & 214 \\ \bottomrule
\end{tabular}}
\end{center}
\end{table}

\paragraph{Dataset Preprocessing.} To filter the trajectories for counterfactual intervention, we preprocess on all the three trials of \{Retail, Airline\}$\times$\{GPT-4o, DeepSeek-V3.1\}. Specifically, we reserve all the tasks that concurrently have \textit{at least} one successful trial in the no-think mode and \textit{at least} one failed trial in the enforced-think mode. Each item in the processed data has a unique failed enforced-think trajectory and its corresponding successful ``golden'' trajectory, which could be shared by multiple failed trials. Finally, we filter out data items whose neither failed nor successful trajectories include any environment-edit actions. This final result is presented as the experimental dataset, as shown in the ``Filter Links'' column in Table~\ref{tab:filter-data}.

\paragraph{Evaluation Metrics.}
We evaluate counterfactual interventions on failed trajectories that contain at least one environment-edit action. Let $\mathcal{T}_{\mathrm{fail}}$ denote the set of such failed trajectories, and let $\tilde{\tau}$ be the intervened trajectory obtained from an original failed trajectory $\tau$. \textbf{First Edit Correction Rate (FECR)} measures the fraction of intervened trajectories whose first environment-edit action is corrected to match the corresponding golden trajectory, among trajectories whose original first edit does not match the golden first edit:
\[
\mathrm{FECR}
=
\frac{
\sum_{\tau \in \mathcal{T}_{\mathrm{fail}}}
\mathbf{1}\!\left[
e_1(\tau) \neq e_1(\tau^{\star})
\;\land\;
e_1(\tilde{\tau}) = e_1(\tau^{\star})
\right]
}{
\sum_{\tau \in \mathcal{T}_{\mathrm{fail}}}
\mathbf{1}\!\left[
e_1(\tau) \neq e_1(\tau^{\star})
\right]
},
\]
where $e_1(\cdot)$ denotes the first environment-edit action and $\tau^{\star}$ denotes the matched golden trajectory.

\textbf{Recovery Rate (RR)} measures the fraction of originally failed trajectories whose final outcome becomes successful after intervention, i.e., matches the golden outcome:
\[
\mathrm{RR}
=
\frac{
\sum_{\tau \in \mathcal{T}_{\mathrm{fail}}}
\mathbf{1}\!\left[
y(\tau) \neq y(\tau^{\star})
\;\land\;
y(\tilde{\tau}) = y(\tau^{\star})
\right]
}{
\sum_{\tau \in \mathcal{T}_{\mathrm{fail}}}
\mathbf{1}\!\left[
y(\tau) \neq y(\tau^{\star})
\right]
},
\]
where $y(\cdot)$ denotes the final task outcome.

\subsection{Turn-level Intervention of Information Disclosure (RQ1)}
\begin{table}[t]
\caption{Counterfactual intervention results (FECR (\%) and RR (\%)) on failure enforce-think agents. Values after `$\pm$' denotes standard deviation.}
\label{tab:counterfactual-main}
\begin{center}
\resizebox{\linewidth}{!}{
\begin{tabular}{lrrrrrrrr}
\toprule
\multirow{3}{*}{\textbf{Intervention}} & \multicolumn{4}{c}{\textbf{Retail}} & \multicolumn{4}{c}{\textbf{Airline}} \\
\cmidrule(lr){2-5} \cmidrule(lr){6-9}
 & \multicolumn{2}{c}{\textbf{GPT-4o}} & \multicolumn{2}{c}{\textbf{DeepSeek-V3.1}} & \multicolumn{2}{c}{\textbf{GPT-4o}} & \multicolumn{2}{c}{\textbf{DeepSeek-V3.1}} \\
 & \textbf{FECR$\uparrow$} & \textbf{RR$\uparrow$} & \textbf{FECR$\uparrow$} & \textbf{RR$\uparrow$} & \textbf{FECR$\uparrow$} & \textbf{RR$\uparrow$} & \textbf{FECR$\uparrow$} & \textbf{RR$\uparrow$} \\ \midrule
Original & 0.00 & 0.00 & 0.00 & 0.00 & 0.00 & 0.00 & 0.00 & 0.00 \\
+Verbosity & 7.09 {\small $\pm$2.65} & 21.35 {\small $\pm$13.64} & 8.33 {\small $\pm$1.07} & {25.93 {\small $\pm$1.50}} & 4.90 {\small $\pm$1.39} & 19.70 {\small $\pm$1.07} & 2.94 {\small $\pm$2.40} & 4.65 {\small $\pm$3.29} \\
+Think Content & 8.51 {\small $\pm$0.00} & 21.88 {\small $\pm$7.76} & 8.33 {\small $\pm$3.86} & 15.34 {\small $\pm$5.24} & \textbf{6.86 {\small $\pm$3.67}} & 18.18 {\small $\pm$1.86} & 0.98 {\small $\pm$1.39} & 3.10 {\small $\pm$1.10} \\
\textbf{+Golden Info} & \textbf{17.73 {\small $\pm$4.01}} & \textbf{33.85 {\small $\pm$0.74}} & {11.36 {\small $\pm$3.21}} & 11.64 {\small $\pm$3.26} & \textbf{6.86 {\small $\pm$1.39}} & \textbf{21.21 {\small $\pm$3.86}} & {8.82 {\small $\pm$2.40}} & {11.63 {\small $\pm$1.90}} \\
\textbf{+Golden Think} & \textbf{17.73 {\small $\pm$4.03}} & {27.60 {\small $\pm$0.90}} & \textbf{28.03 {\small $\pm$1.31}} &  \textbf{32.28\small $\pm$1.83} & {5.88 {\small $\pm$2.94}} & {17.42 {\small $\pm$3.47}} & \textbf{9.80 {\small $\pm$4.49}} & \textbf{20.93 {\small $\pm$}6.15} \\ \bottomrule
\end{tabular}}
\end{center}
\end{table}

\par We first examine the causal effect of different turn-level information injection methods on failed enforced-think trajectories. We aim to answer the following specific questions under RQ1: 
\begin{itemize}
    \item RQ 1.1: Given an actually observed failed enforced-think trajectory, is exposing the private reasoning generated in that failed trajectory sufficient to recover the missing communication and change the subsequent interaction?
    \item RQ 1.2: Can we change the subsequent user decision and agent action if we restore the specific decision-relevant information identified as missing from the failed response?
    \item RQ 1.3: If we rerun the same task until the thinking agent succeeds, can the reasoning trace from that successful run contain information useful for repairing the failed interaction?
\end{itemize}

We design four intervention groups to address the RQs above: \texttt{+Verbosity} increases the response length without adding missing golden evidence, serving as a baseline. \texttt{+Think Content} injects information extracted from the agent's private think trace to address RQ 1.1. \texttt{+Golden Info} injects missing decision-relevant information extracted from a matched successful, non-thinking trajectory to address RQ 1.2. \texttt{+Golden Think} injects information extracted from the agent's reasoning content from a matched successful, thinking trajectory to address RQ 1.3.

\begin{figure*}[t]
\centering

\begin{minipage}[t]{0.42\linewidth}
\centering
\vspace{0pt}                       
\captionof{table}{Summary of the effect of interventions on task dynamics micro-averaged on models and domains.}
\label{tab:intervention}
\resizebox{\linewidth}{!}{\begin{tabular}{lrr}
\toprule
\textbf{Intervention} & \textbf{FECR$\uparrow$ (\%)} & \textbf{RR$\uparrow$ (\%)} \\ \midrule
Original       & 0.00           & 0.00           \\
+Verbosity     & 6.08 {\small $\pm$0.30}  & 19.00 {\small $\pm$2.96} \\
+Think Content & 6.50 {\small $\pm$1.57}  & 15.42 {\small $\pm$2.50} \\
\textbf{+Golden Info}   & {11.74 {\small $\pm$1.57}} & {20.25 {\small $\pm$1.17}} \\
\textbf{+Golden Think}   & \textbf{16.35 {\small $\pm$2.05}} & \textbf{25.55 {\small $\pm$2.10}} \\\bottomrule
\end{tabular}}
\end{minipage}
\hfill
\begin{minipage}[t]{0.56\linewidth}
\centering
\vspace{0pt}
\captionof{table}{Ratio of the number of trials to obtain a successful reasoning trajectory to construct the \texttt{+Golden Think} instance group.}\label{tab:success-trials}
\resizebox{\linewidth}{!}{\begin{tabular}{llrrr}
\toprule
\textbf{Model} & \textbf{Domain} & \textbf{1 trials} & \textbf{2 trials} & \textbf{3 trials} \\
\midrule
\multirow{2}{*}{GPT-4o} & Retail & 67.86\% & 21.43\% & 10.71\% \\
 & Airline & 68.75\% & 25.00\% & 6.25\% \\
\multirow{2}{*}{DeepSeek-V3.1} & Retail & 62.50\% & 29.17\% & 8.33\% \\
 & Airline & 83.33\% & 11.11\% & 5.56\% \\ \bottomrule
\end{tabular}}
\end{minipage}
\end{figure*}

 \begin{figure}
\centering
    \includegraphics[width=0.6\linewidth]{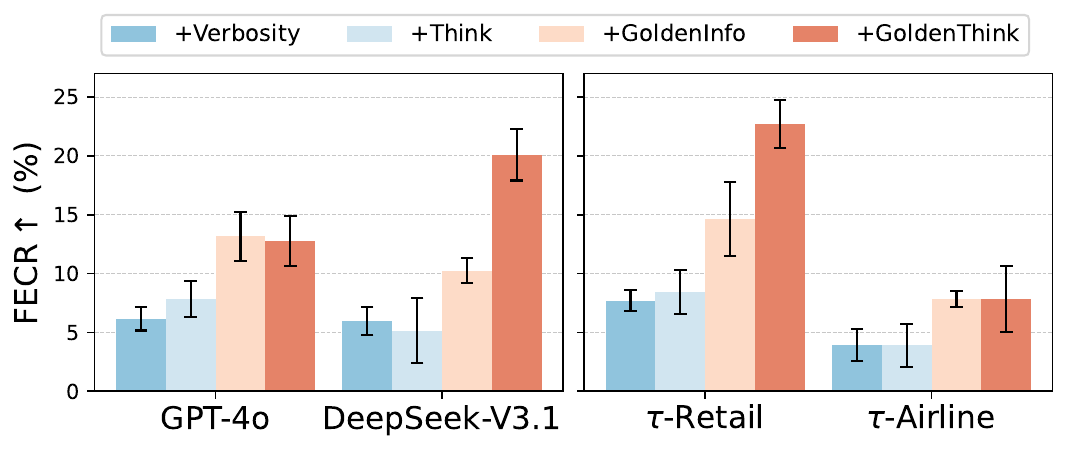}
\caption{Left: Intervention effect w.r.t. models. Right: Intervention effect w.r.t. domains. Error bars denotes standard deviation.}
\label{fig:counterfactual-summary}
\end{figure}

\par Table~\ref{tab:counterfactual-main} reports the results across two models and two domains. Table~\ref{tab:intervention} and  Figure~\ref{fig:counterfactual-summary} summarizes the micro-averaged intervention effects. Overall, \texttt{+Golden Think} produces the clearest effect on the agent's immediate downstream action. Its micro-averaged FECR reaches $16.35\%$, compared with $6.08\%$ for \texttt{+Verbosity} and $6.50\%$ for \texttt{+Think Content}, and the same trend holds under both model-level and domain-level aggregation in Figure~\ref{fig:counterfactual-summary}. \texttt{+Golden Info} also exhibits significant improvment compared to baseline interventions. Both interventions demonstrate the utility of injecting missing information to failed trajectories. Since FECR measures whether the first environment-edit action is corrected after intervention, this result suggests that missing decision-relevant disclosure can causally affect the agent's immediate structured behavior after the user decision point. On the other hand, the weaker effect on final recovery indicates that RR is a more downstream and noisier outcome, as later turns, tool calls, or planning decisions may still fail even after the first edit is corrected. 

It also shows that internal thinking is counter-productive for the interactive multi-decision setting. The \texttt{+Think Content} does not consistently outperform \texttt{+Verbosity}, whereas \texttt{+Golden Info} more reliably improves FECR.  \texttt{+Golden Think} requires multiple trials obtain a successful trajectory (see Table~\ref{tab:success-trials}), although it presents a significant improvement of FECR. Thus, targeted information disclosure has a clearer causal effect on immediate action correction than generic verbosity or direct think-content injection, while its effect on end-to-end recovery remains setting-dependent.

\subsection{Intervention Frequency of Different Dimensions (RQ2)}

\begin{figure}[t]
    \centering
    \begin{subfigure}{0.49\textwidth}
        \centering
        \includegraphics[width=\linewidth]{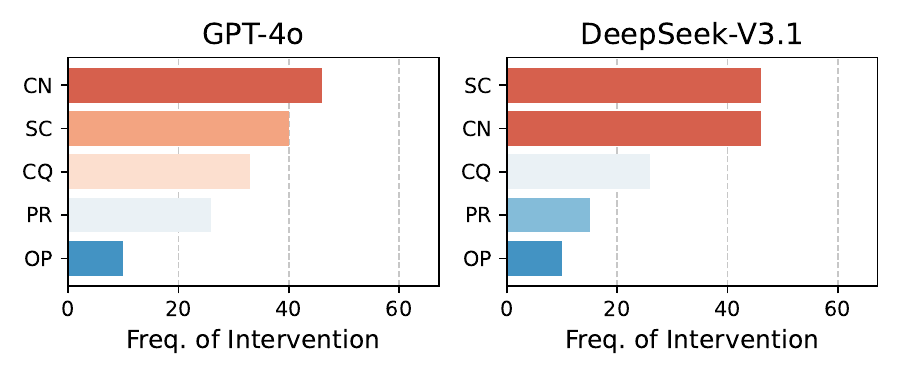}
        \caption{Frequency of intervention on different models.}
        \label{fig:counterfactual-dimension-model}
    \end{subfigure}
    \hfill
    \begin{subfigure}{0.49\textwidth}
        \centering
        \includegraphics[width=\linewidth]{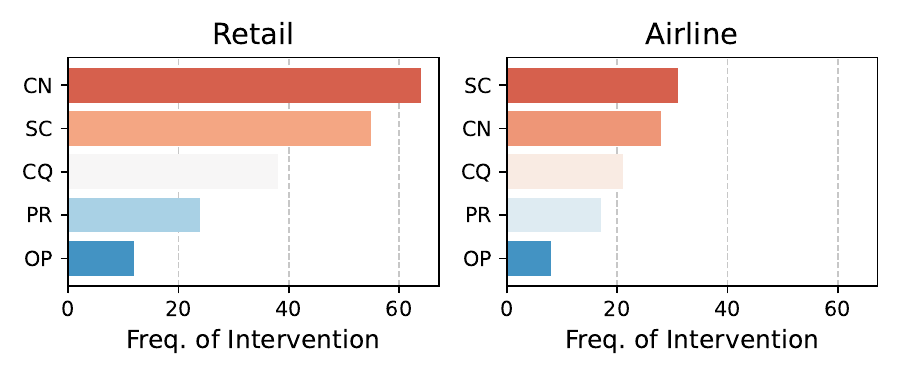}
        \caption{Frequency of intervention on different scenarios.}
        \label{fig:counterfactual-dimension-env}
    \end{subfigure}
    \caption{Results of frequency of intervention for different information disclosure dimensions.}
    \label{fig:infodis-frequency}
\end{figure}

\begin{table}[t]
\vspace{-10pt}
    \centering
    \caption{Task performance of agents with various prompting across three independent trials. Each value is displayed as a percentage. Values in parentheses denote the change against Baseline.}\label{tab:infodis-prompt}
\resizebox{\linewidth}{!}{\begin{tabular}{ll|rrrr|rr}
\toprule
\multirow{2}{*}{\textbf{Model}} & \multirow{2}{*}{\textbf{Prompt}} & \multicolumn{2}{c}{\textbf{Retail}} & \multicolumn{2}{c|}{\textbf{Airline}} & \multicolumn{2}{c}{\textbf{Micro Overall}} \\
 &  & \textbf{Pass\^{}1}  & \textbf{Pass\^{}2}  & \textbf{Pass\^{}1}  & \textbf{Pass\^{}2}   & \textbf{Pass\^{}1}  & \textbf{$\Delta$(Pass\^{}1)} \\
 \midrule
 \rowcolor{gray!25} \multicolumn{8}{c}{{tool-calling agent w/o enforced think tool}} \\ 
\multirow{3}{*}{GPT-4o} & Baseline & 64.06 & 53.33 & 43.33 & {30.00} & 57.78 & - \\
 & ProCoT & 66.38 {\small (+2.32)} & 55.07 {\small (+1.74)} & 46.00 {\small (+2.67)} & 32.00 {\small (+2.00)} & 60.00 & +2.42\\
 & InfoDis (ours) & {67.25 {\small (+3.19)}} & {54.49 {\small (+1.16)}} & {45.33 {\small (+2.00)}} & 27.33 {\small (-2.67)}  & {60.61} & \textbf{+2.83} \\ 
 \cmidrule(lr){1-6} \cmidrule(lr){7-8}
 \multirow{3}{*}{DeepSeek-V3.1} & Baseline & {71.01} & {59.71} & 50.00 & 34.67 & 64.65 & - \\
 & ProCoT & 70.14 {\small (-0.87)} & 60.00 {\small (+0.29)} & 51.33 {\small (+1.33)} & 37.33 {\small (+2.66)} & 64.44 & -0.21 \\
 & InfoDis (ours) & 70.43 {\small (-0.58)} & 57.97 {\small (-1.74)} & {52.00 {\small (+2.00)}} & {36.67 {\small (+2.00)}} & {64.85} & \textbf{+0.20} \\
 \rowcolor{gray!25} \multicolumn{8}{c}{{tool-calling agent w/ enforced think tool}} \\ 
\multirow{2}{*}{GPT-4o} & Baseline & 59.71 & 45.51 & 35.33 & 19.33 & 52.32 & - \\
 & InfoDis (ours) & {61.74 {\small (+2.03)}} & {49.57 {\small (+4.06)}} & {48.00 {\small (+12.67)}} & {33.33 {\small (+14.00})} & {57.58} & \textbf{+5.26} \\  
 \cmidrule(lr){1-6} \cmidrule(lr){7-8}
\multirow{2}{*}{DeepSeek-V3.1} & Baseline & 63.48 & 46.09 & {49.33} & 35.33 & 59.19 & - \\
 & InfoDis (ours) & {68.70 {\small (+5.22)}} & {54.20 {\small (+8.11)}} &  48.67 {\small (-0.66)} & {36.00 {\small (+0.67)}} & {62.63} & \textbf{+3.44} \\ \bottomrule
\end{tabular}}
\end{table}

\par Figure~\ref{fig:infodis-frequency} shows the frequency of InfoDis categories injected by \texttt{+Golden Info}, corresponding to information present in matched successful trajectories but absent from failed target trajectories. Across both model-level and domain-level views, \textit{State Correction (SC)} and \textit{Constraint (CN)} are the most frequent missing categories, whereas \textit{Option (OP)} is consistently the least frequent. This pattern supports our hypothesis that failed trajectories often lack information that directly affects the user's next decision: state correction updates the user's understanding of the current task state, while constraint disclosure informs the user about rules and limits on valid actions.

\par This result also aligns with the motivating case study in Figure~\ref{fig:case-study}, where the enforced-think agent omits a cancellation constraint before asking the user for confirmation. The aggregate analysis shows that such missing constraints are not isolated; together with state corrections, they form the largest gap between failed and successful trajectories. Therefore, both qualitative and quantitative evidence suggest that user-service failures are often associated with missing decision-relevant disclosure, especially information that changes the user's understanding of the current state or the validity of possible actions.

\subsection{Prompt-level Intervention of Information Disclosure (RQ3)}

\par Finally, we evaluate whether improving information disclosure at the prompt level can boost agent performance. We add a lightweight InfoDis instruction to the end of the agent system prompt (see Appendix \ref{app:infodis-prompt}), encouraging the agent to disclose decision-relevant states, constraints, options, and consequences before asking the user to make decisions. We compare InfoDis against two prompt strategies: Baseline and ProCoT~\citep{deng2023prompting}. Baseline is the original prompting of the tool-calling agent in $\tau$-bench, including task policy and action guidelines. ProCoT prompts the model to reason about the dialogue state and plan a proactive next action before generating its response, making it a relevant comparison for InfoDis. Unlike the turn-level intervention in RQ1, this intervention is applied before the trajectory starts and is evaluated on the full Retail and Airline test sets across three independent trials.

\par As shown in Table~\ref{tab:infodis-prompt}, InfoDis prompting generally recovers task performance. For agents without the think tool, it further raises the micro-averaged pass\^{}1 of GPT-4o from $57.78\%$ to $60.61\%$, while maintaining a slight gain for DeepSeek-V3.1. More importantly, for enforced-think agents, InfoDis prompting consistently mitigates the performance degradation. GPT-4o improves from $52.32\%$ to $57.58\%$ micro pass\^{}1, and DeepSeek-V3.1 improves from $59.19\%$ to $62.63\%$. The gains are especially clear on Retail, where enforced-think agents improve on both pass\^{}1 and pass\^{}2 for both models.

\par These results suggest that information disclosure is not only a diagnostic lens for analyzing failed trajectories, but also a practical intervention target. For already strong no-think agents, InfoDis prompting can raise the performance ceiling by improving user-facing communication. For enforced-think agents, it helps recover part of the lost performance by encouraging public response to be externalized as decision-relevant disclosure. This supports our central claim that user-service agents should not only reason privately, but also communicate the right information before user decision points.

\section{Conclusion}
\label{sec:conclusion}

\par In this work, we rethink user-engaged LLM agents from the perspective of agent-user communication. Across multiple models and user-service scenarios, we find that enforced internal thinking does not reliably mitigate task performance, and increasing thinking effort does not necessarily translate to user-facing communication. 
To explain this gap, we formulate \textit{information disclosure} as a communication-level mechanism in agent-user interaction. We define and probe five types of decision-relevant disclosure: state correction, option, constraint, consequence, and proactive retrieval. Through evidence-grounded annotation and turn-level counterfactual intervention, we show that failed trajectories often lack information that is present in matched successful trajectories. Injecting golden decision-relevant information produces the clearest improvement in First Edit Correction Rate, indicating that missing disclosure can causally affect the agent's immediate downstream environment-edit behaviour. At the same time, the more modest gains in final Recovery Rate show that information disclosure explains an actionable subset of failures rather than all failures. Our results also show that raw internal thinking is not equivalent to effective user-facing disclosure. Directly injecting think content is less reliable than injecting targeted golden information, suggesting that the current usage of the think tool does not explicitly boost the models' capability of capturing pivotal information for disclosures. Finally, our prompt-level InfoDis intervention improves performance for no-think agents and mitigates the degradation of enforced-think agents, demonstrating that information disclosure is not only a diagnostic lens but also a practical design target.

\par Overall, our findings suggest that future user-service agents should be optimized not merely to enhance reasoning ability, but to improve communicating the right information at the right decision points. The central challenge of user-service agents is therefore the reliable conversion of environment information into public, decision-relevant disclosure. We hope this study could incentivise future research in user-centric agents and user-agent interaction.

\subsubsection*{Acknowledgements}
\par The authors would like to thank Seongheon Park and Leitian Tao for their valuable suggestions on the manuscript. This work is supported in part by the AFOSR Young Investigator Program under award number FA9550-23-1-0184, Office of Naval Research 
under award N000142612508, National Science Foundation under awards IIS-2237037 and IIS-2331669,  Schmidt Sciences Foundation, Open Philanthropy (now Coefficient Giving), Alfred P. Sloan Fellowship, and gifts from Google and Amazon.

\bibliography{main}
\bibliographystyle{tmlr}

\newpage

\clearpage
\appendix
\section*{Appendix Contents}

\noindent
\hyperref[app:prompt]
    {\textbf{Appendix A: Prompts Used in Experiments}}
\dotfill \pageref{app:prompt}

\medskip

\noindent
\hyperref[app:think-focus]
    {\textbf{Appendix B: Performance Comparison Under Different Think Focus and User Simulator}}
\dotfill \pageref{app:think-focus}

\medskip

\noindent
\hyperref[app:infodis-config]
    {\textbf{Appendix C: Configurations of Evidence-grounded Probing and Information Injection}}
\dotfill \pageref{app:infodis-config}

\medskip

\noindent
\hyperref[app:check-infodis]
    {\textbf{Appendix D: Consistency Check of Evidence-Grounded Probing}}
\dotfill \pageref{app:check-infodis}

\medskip

\noindent
\hyperref[app:case-study]
    {\textbf{Appendix E: Case Study of Counterfactual Intervention}}
\dotfill \pageref{app:case-study}

\section{Prompts Used in Experiments}\label{app:prompt}
\subsection{Think Effort Control}

\begin{promptbox}{Low Think (System-1) Prompt}
You must use System 1 thinking. Before providing your final answer, provide a direct, concise, and intuitive answer immediately. Do not engage in step-by-step reasoning, internal monologue, or detailed deliberation. Focus on speed and immediate pattern recognition.
\end{promptbox}

\begin{promptbox}{High Think (System-2) Prompt}
You must use System 2 thinking. Before providing your final answer, engage in an extensive, step-by-step analytical process. Deconstruct the problem, evaluate potential pitfalls, verify your logic, and show your detailed reasoning. Prioritize depth, accuracy, and rigorous deliberation over brevity.
\end{promptbox}

\subsection{Evidence-grounded Information Probing}

\begin{promptbox}{Evidence-Grounded Information Probing Prompt}
You are an expert annotator for information disclosure in user-agent task-oriented dialogues. Your job is to analyze each assistant turn in a conversation and measure how well the agent discloses decision-relevant information to the user.

## Your Task

You will receive a full conversation as a JSON array. Each object has:
- "index": turn number (0-indexed)
- "role": "user" or "assistant"  
- "content": the turn text (may be null)

For each assistant turn where content is NOT null, you must annotate the information disclosure across five categories. For each category you must work in two steps, in this exact order:

1. **Mark the evidence first.** Find the verbatim span of text in the SAME assistant turn that discloses the information for this category, and copy it exactly into the "quote" field. The quote MUST be an exact substring of that assistant turn's content --- do not paraphrase, summarize, normalize whitespace, or stitch together non-contiguous fragments. If multiple spans are relevant, choose the single most decision-relevant contiguous span.
2. **Then assign the binary label.** Decide "present" or "absent" based on whether a valid supporting quote exists.

The label MUST be justified by the quote and nothing else. If you cannot point to an exact span in the current turn that discloses the information, the label is "absent" and the "quote" field MUST be an empty string ("").

Skip turns where content is null or the role is "user".

## Information Categories

Evaluate each category only with respect to the USER'S UPCOMING DECISION at that point in the conversation. A piece of information is "decision-relevant" if it would change or enable the user's next action. Information the user already provided or could infer from prior context does NOT count.

### 1. State-Correction (SC)
Information that corrects a factual misunderstanding or false assumption the user has expressed.
- Present: a span in the turn explicitly corrects the user's mistaken belief (e.g., user says "my cancelled order" but agent clarifies "this order was delivered, not cancelled")
- Absent: agent doesn't correct a visible user misconception and proceeds as if it were true

### 2. Option (OP)
The set of choices available to the user for their upcoming decision --- including relevant variants, alternatives, or paths.
- Present: a span in the turn lists available options for the user to choose
- Absent: options exist but are not disclosed, or only one option is mentioned without indication of whether others exist

### 3. Constraint (CN)
Rules, policies, eligibility conditions, or limits that affect what the user can or cannot do --- and that the user likely does not already know.
- Present: a span in the turn explicitly states the relevant policy or constraint (e.g., "this order is eligible for return because it has been delivered", "modifications are only possible before shipment")
- Absent: a constraint available for the agent and can affects the user's decision, but is not disclosed

### 4. Consequence (CQ)
The concrete outcomes that will result from the user's decision --- including financial impact (amounts, not just method), timing, and irreversible effects.
- Present: a span in the turn discloses material consequences (e.g., refund amount, refund timeline, charge amount, order status change)
- Absent: consequences exist but are not disclosed at a decision point

### 5. Proactive Retrieval (PR)
Information the agent proactively fetched from its tools/database and disclosed to the user --- information the user did not possess and did not directly ask for, but which is necessary or highly useful for the user's decision.
- Present: a span in the turn discloses retrieved user-specific data that enables the decision (e.g., listing the user's orders with their contents, showing both current and new address for comparison, showing current item specs alongside alternatives)
- Absent: agent could have retrieved and disclosed helpful user-specific data but did not

## Scoring Labels

For each category, use exactly one of:
- "present" --- fully and clearly disclosed, supported by a verbatim quote from the turn
- "absent" --- should have been disclosed but was not; quote is an empty string

## Critical Evaluation Rules

1. **Quote before label**: Always identify the supporting verbatim span before deciding the label. A "present" label without a valid exact-substring quote is invalid. An "absent" label must have an empty-string quote.

2. **No fabricated or paraphrased quotes**: The "quote" value must appear character-for-character in the current assistant turn's content. Never invent, paraphrase, or pull text from a different turn.

3. **Context carries forward**: If information was disclosed in an earlier turn and is still valid, do not mark it present again from a later turn unless that later turn itself restates it (quotes are always drawn from the current turn).

4. **Null content turns**: If content is null, output a null entry for that turn index --- do not attempt to annotate.

5. **Do not infer what the agent "meant"**: Judge only what is explicitly stated in a quotable span. Implied or inferable information should be marked "absent" with an empty-string quote.

## Output Format

Return a JSON array. Each element corresponds to one assistant turn (non-null content only). Null turns must still appear in the output with null annotation fields. Do not include user turns in your output.

```json
[
  {
    "turn_index": <int>,
    "role": "assistant",
    "is_null": <true|false>,
    "decision_point": <true|false>,
    "annotations": {
      "SC": { "quote": "<exact substring of this turn, or \"\" if absent>", "label": "<present|absent>" },
      "OP": { "quote": "<exact substring of this turn, or \"\" if absent>", "label": "<present|absent>" },
      "CN": { "quote": "<exact substring of this turn, or \"\" if absent>", "label": "<present|absent>" },
      "CQ": { "quote": "<exact substring of this turn, or \"\" if absent>", "label": "<present|absent>" },
      "PR": { "quote": "<exact substring of this turn, or \"\" if absent>", "label": "<present|absent>" }
    }
  }
]
```

For null-content turns, output:

```json
{ "turn_index": <int>, "role": "assistant", "is_null": true, "decision_point": false, "annotations": null }
```

The `decision_point` field is true if this turn ends with an explicit or implicit request for a user decision (confirmation, selection, or provision of information).
\end{promptbox}

\subsection{Information Injection for Counterfactual Intervention}

\begin{promptbox}{Golden Information Injection - System Prompt}
You are rewriting a customer-service agent's response to include additional information that the agent should have disclosed to the user. Your goal is to produce a single, natural-sounding message that keeps the original response's intent and tone while seamlessly integrating the missing information listed below.

Rules:
- Preserve all original information; do not remove anything.
- Integrate each piece of missing information naturally into the message.
- Do not add information beyond what is explicitly listed as missing.
- Do not mention that the response was enhanced or modified.
- Keep a professional customer-service tone.
  - Output ONLY the enhanced response text, nothing else.
\end{promptbox}

\begin{promptbox}{Golden Information Injection - User Prompt}
## Conversation so far (for context):
{conversation_context}

## Agent's original response at the current turn:
{original_response}

## Missing information to incorporate:
- **{category} ({description})**: {quote}
...

## Enhanced response:
\end{promptbox}

\begin{promptbox}{Think Content Injection - System Prompt}
You are rewriting a customer-service agent's response to naturally incorporate the agent's internal reasoning into the message. The agent had internal thoughts before composing its response; your job is to weave the relevant information from those thoughts into the response so the user benefits from the reasoning.

Rules:
- Preserve all original response information; do not remove anything.
- Integrate relevant details from the agent's internal thoughts naturally.
- Do not add information beyond what is in the thoughts and the original response.
- Do not mention that the response was enhanced, modified, or that there were internal thoughts
- Keep a professional customer-service tone.
- Output ONLY the enhanced response text, nothing else.
\end{promptbox}

\begin{promptbox}{Think Content Injection - User Prompt}
## Agent's internal thoughts (private, not shown to user):
{think_content}

## Agent's original response to the user:
{original_response}

## Enhanced response (incorporating relevant details from the thoughts):
\end{promptbox}

\begin{promptbox}{Verbosity Injection - System Prompt}
You are rewriting a customer-service agent's response to be more verbose and detailed. Expand the response with more elaborate phrasing, additional politeness, and fuller explanations of the same points already made.

Rules:
- Do NOT add any new factual information, options, or details that are not
already present in the original response.
- Only expand on what is already stated: use more words, more polite phrasing,
restate key points for emphasis, and add transitional phrases.
- Maintain the same professional customer-service tone.
- The response should be noticeably longer than the original.
- Do not mention that the response was rewritten or enhanced.
- Output ONLY the verbose response text, nothing else.
\end{promptbox}

\begin{promptbox}{Verbosity Injection - User Prompt}
## Agent's original response:
{original_response}

## More verbose version of the same response (no new information):
\end{promptbox}

\subsection{Information Disclosure Prompting}\label{app:infodis-prompt}
The information disclosure prompt is appended to the end of the system prompt of user-service agents, as shown below.

\begin{promptbox}{Information Disclosure -- System Prompt}
## Information Disclosure
- You must interact with the user actively and disclose as much information as possible to ensure the user is well-informed about the current state of the database and any potential changes.    
\end{promptbox}
\section{Performance Comparison Under Different Think Focus and User Simulator}\label{app:think-focus}

Table~\ref{tab:think-focus} further examines whether the instability of enforced thinking can be explained by the user simulator or by the specific instruction of the think tool. First, we replace the GPT-4.1 user simulator with a stronger GPT-5 simulator while keeping GPT-4o as the agent. The results remain mixed: compared with the no-think baseline, the default enforced-think setting still reduces micro-average performance from $56.16\%$ to $51.31\%$ in pass\^{}1 and from $43.23\%$ to $39.39\%$ in pass\^{}2. This suggests that the unstable utility of the think tool is not merely caused by an insufficiently capable user simulator; rather, the limitation mainly arises from the agent side.

\par Second, we modify the description of the think tool to encourage different reasoning focuses. An action-centric think tool emphasizes planning the next tool call or environment action, while a user-centric think tool emphasizes the user's information needs and decision context. However, the effect of these variants is not consistent across models and user simulators. For GPT-4o with the GPT-4.1 user simulator, the user-centric focus improves the micro-average pass\^{}1 and pass\^{}2 over the action-centric focus. In contrast, for DeepSeek-V3.1, the action-centric focus performs better than the user-centric focus. With the GPT-5 user simulator, both focused variants improve over the default think setting, but neither consistently recovers the no-think baseline. These results indicate that different models respond differently to the same think-tool instruction, likely because their internal reasoning and tool-use behaviors differ. Therefore, simply tuning the description or focus of the think tool is unlikely to provide a general solution to the instability of enforced internal thinking.

\begin{table}[t]
    \centering
    \caption{Performance comparison of different think focus.}
    \label{tab:think-focus}
    \resizebox{\linewidth}{!}{\begin{tabular}{ll|rrrr|rr}
\toprule
\multirow{2}{*}{Model}& \multirow{2}{*}{Think Focus} & \multicolumn{2}{c}{\textbf{Retail}} & \multicolumn{2}{c|}{\textbf{Airline}} & \multicolumn{2}{c}{\textbf{Micro Overall}} \\
 &  & \textbf{Pass\^{}1} & \textbf{Pass\^{}2} & \textbf{Pass\^{}1} & \textbf{Pass\^{}2} & \textbf{Pass\^{}1} & \textbf{Pass\^{}2} \\
\midrule
\multirow{2}{*}{\shortstack[l]{Agent: GPT-4o\\User: GPT-4.1}}
 & Action-centric & 56.81 & 45.22 & 43.33 & 26.67 & 52.73 & 39.60 \\
 & User-centric   & 61.74 & 50.14 & 41.33 & 24.67 & 55.56 & 42.42 \\
\midrule
\multirow{4}{*}{\shortstack[l]{Agent: GPT-4o\\User: GPT-5}}
 & w/o think      & 63.48 & 50.72 & 39.33 & 26.00 & 56.16 & 43.23 \\
 & Default        & 57.97 & 46.09 & 36.00 & 24.00 & 51.31 & 39.39 \\
 & Action-centric & 59.42 & 49.86 & 36.00 & 20.67 & 52.32 & 41.01 \\
 & User-centric   & 60.87 & 49.28 & 35.33 & 23.33 & 53.13 & 41.41 \\
\midrule
\multirow{2}{*}{\shortstack[l]{Agent: DeepSeek-V3.1\\User: GPT-4.1}}
 & Action-centric & 66.38 & 52.75 & 50.67 & 33.33 & 61.62 & 46.87 \\
 & User-centric   & 64.64 & 50.14 & 50.00 & 32.67 & 60.20 & 44.85 \\
\bottomrule
\end{tabular}}
\end{table}

\begin{promptbox}{Default Think Tool - Configuration}
{
    "type": "function",
    "function": {
        "name": "think",
        "description": (
            "Use the tool to think about something. It will not obtain new information or change the database, "
            "but just append the thought to the log. Use it when complex reasoning or some cache memory is needed."
        ),
        "parameters": {
            "type": "object",
            "properties": {
                "thought": {
                    "type": "string",
                    "description": "A thought to think about.",
                },
            },
            "required": ["thought"],
        },
    },
}
\end{promptbox}

\begin{promptbox}{Action-Centric Think Tool - Configuration}
{
    "type": "function",
    "function": {
        "name": "think",
        "description": (
            "Use the tool to think about something. It will not obtain new information or change the database, "
            "but just append the thought to the log. Reason about the database state, tools, and your next action."
        ),
        "parameters": {
            "type": "object",
            "properties": {
                "thought": {
                    "type": "string",
                    "description": "A thought to think about.",
                },
            },
            "required": ["thought"],
        },
    },
}
\end{promptbox}

\begin{promptbox}{User-Centric Think Tool - Configuration}
{
    "type": "function",
    "function": {
        "name": "think",
        "description": (
            "Use the tool to think about something. It will not obtain new information or change the database, "
            "but just append the thought to the log. Reason about what the user needs to know to make an informed decision."
        ),
        "parameters": {
            "type": "object",
            "properties": {
                "thought": {
                    "type": "string",
                    "description": "A thought to think about.",
                },
            },
            "required": ["thought"],
        },
    },
}
\end{promptbox}
\section{Configurations of Evidence-grounded Probing and Information Injection}\label{app:infodis-config}
\par In the evidence-grounded information probing and information injection, we use GPT-4.1 as the annotator. The \texttt{temperature} is set to 0.2. To avoid format or syntax errors in annotation, we apply a rule-based format check and enable the annotator to retry at most two times when triggering an exception.

\section{Consistency Check of Evidence-Grounded Probing}\label{app:check-infodis}

We repeat the LLM-as-a-judge annotation for two times for annotation w/o evidence and w/ evidence, respectively. As shown in Table~\ref{tab:self-consistency}, evidence-grounded annotation criteria can significantly improve the self-consistency of annotation, improving from 0.70$\sim$0.94 to 0.86$\sim$0.94. For example, without evidence, the annotator's cohen's $\kappa$ on \textit{Constraint} is only 0.7079, which improves to 0.8681 with evidence grounding. As a result, evidence-grounding significantly makes the LLM-as-a-judge annotation more stable. Moreover, the evidence annotated in this stage directly provides the material for counterfactual intervention via information injection.

\begin{table}[t]
\caption{Cohen's $\kappa$ results on self-consistency of LLM-as-a-judge annotation.}\label{tab:self-consistency}
\resizebox{\linewidth}{!}{\begin{tabular}{lllllll}
\toprule
Criteria & State Correction & Option & Constraint & Consequence & Proactive Retrieval & Decision Point \\ \midrule
w/o evidence & 0.8443 & 0.7766 & 0.7079 & 0.8301 & 0.8363 & 0.9308 \\
w/ evidence & \textbf{0.9177} & \textbf{0.8976} & \textbf{0.8681} & \textbf{0.9384} & \textbf{0.9188} & \textbf{0.9362} \\
\bottomrule
\end{tabular}}
\end{table}
\section{Case Studies of Counterfactual Intervention}\label{app:case-study}
\par We present cases of the counterfactual intervention with the three methods: \texttt{+Golden Info}, \texttt{+Think Content} and \texttt{+Verbosity}. In operation, each intervention is indexed by its \texttt{failed\_index} and \texttt{golden\_index}, which share the same \texttt{task\_id} while the \texttt{trial} comes from the failed trajectory. The \texttt{turn\_id} represents the index of the intervened turn in the failed trajectory, while the \texttt{first\_edit\_turn} denotes the turn index where the first environment edit happens. The \texttt{original\_response} denotes the target response in the failed trajectory to be intervened, while the \texttt{enhanced\_response} denotes the response intervened by adding missing information / think content/ verbosity.

\begin{promptbox}{Case: +Golden Info}
{
  "task_id": 27,
  "trial": 2,
  "failed_index": 17,
  "golden_index": 13,
  "turn_id": 24,
  "first_edit_turn": 26,
  "original_response": "Here are the details for your requests:\n\n1. **Return**:\n   - Items to be returned:\n     - Garden Hose (Item ID: 5753502325)\n     - Backpack (Item ID: 9851293632)\n   - Refund will be processed to your PayPal account.\n\n2. **Exchange**:\n   - Item to be exchanged:\n     - Hiking Boots (Item ID: 8118291112)\n   - New item:\n     - Waterproof Hiking Boots, Size 12, Leather (Item ID: 8277474082)\n   - Price difference: The new boots cost $236.57, and the original boots cost $260.56. You will receive a refund of $23.99 to your PayPal account.\n\nPlease confirm if all the details are correct and if you would like me to proceed with these actions.",
  "enhanced_response": "Here are the details for your requests:\n\n1. **Return**:\n   - Items to be returned:\n     - Garden Hose (Item ID: 5753502325)\n     - Backpack (Item ID: 9851293632)\n   - Refund will be processed to your PayPal account.\n\n2. **Exchange**:\n   - Item to be exchanged:\n     - Hiking Boots (Item ID: 8118291112)\n   - New item:\n     - Waterproof Hiking Boots, Size 12, Leather (Item ID: 8277474082)\n   - Price difference: The new boots cost $236.57, and the original boots cost $260.56. You will receive a refund of $23.99 to your PayPal account.\n\nUnfortunately, I cannot process both a return and an exchange for the same order. I can only process one type of request at a time. Please confirm if all the details are correct and let me know which request you would like me to proceed with.",
  "missing_categories": [
    "SC",
    "CN"
  ],
  "missing_info": {
    "SC": "Unfortunately, I cannot process both a return and an exchange for the same order.",
    "CN": "I can only process one type of request at a time."
  }
}
\end{promptbox}

\begin{promptbox}{Case: +Think Content}
{
  "task_id": 27,
  "trial": 2,
  "failed_index": 17,
  "turn_id": 24,
  "first_edit_turn": 26,
  "original_response": "Here are the details for your requests:\n\n1. **Return**:\n   - Items to be returned:\n     - Garden Hose (Item ID: 5753502325)\n     - Backpack (Item ID: 9851293632)\n   - Refund will be processed to your PayPal account.\n\n2. **Exchange**:\n   - Item to be exchanged:\n     - Hiking Boots (Item ID: 8118291112)\n   - New item:\n     - Waterproof Hiking Boots, Size 12, Leather (Item ID: 8277474082)\n   - Price difference: The new boots cost $236.57, and the original boots cost $260.56. You will receive a refund of $23.99 to your PayPal account.\n\nPlease confirm if all the details are correct and if you would like me to proceed with these actions.",
  "enhanced_response": "Here are the details for your requests regarding order #W7181492:\n\n1. **Return**:\n   - Items to be returned:\n     - Garden Hose (Item ID: 5753502325)\n     - Backpack (Item ID: 9851293632)\n   - Refund will be processed to your PayPal account.\n2. **Exchange**:\n   - Item to be exchanged:\n     - Hiking Boots (Item ID: 8118291112)\n   - New item:\n     - Waterproof Hiking Boots, Size 12, Leather (Item ID: 8277474082), which is available.\n   - Price difference: The new boots cost $236.57, and the original boots cost $260.56. You will receive a refund of $23.99 to your PayPal account.\n\nPlease confirm if all the details are correct and if you would like me to proceed with these actions.",
  "think_content": "The hose and backpack are part of order #W7181492, and the hiking boots are also part of the same order. The waterproof variant of the hiking boots in size 12 is available with item ID 8277474082. I will now prepare the return and exchange details for confirmation."
}
\end{promptbox}

\begin{promptbox}{Case: +Verbosity}
{
  "task_id": 27,
  "trial": 2,
  "failed_index": 17,
  "turn_id": 24,
  "first_edit_turn": 26,
  "original_response": "Here are the details for your requests:\n\n1. **Return**:\n   - Items to be returned:\n     - Garden Hose (Item ID: 5753502325)\n     - Backpack (Item ID: 9851293632)\n   - Refund will be processed to your PayPal account.\n\n2. **Exchange**:\n   - Item to be exchanged:\n     - Hiking Boots (Item ID: 8118291112)\n   - New item:\n     - Waterproof Hiking Boots, Size 12, Leather (Item ID: 8277474082)\n   - Price difference: The new boots cost $236.57, and the original boots cost $260.56. You will receive a refund of $23.99 to your PayPal account.\n\nPlease confirm if all the details are correct and if you would like me to proceed with these actions.",
  "enhanced_response": "Thank you very much for reaching out with your requests. I have carefully reviewed the details you provided, and I would like to summarize them here for your confirmation before proceeding further:\n\n1. **Return Request**:\n   - The following items are scheduled to be returned:\n     - Garden Hose, with the Item ID: 5753502325\n     - Backpack, with the Item ID: 9851293632\n   - Once these items are returned, the refund for both will be processed directly to your PayPal account, ensuring a seamless and convenient transaction for you.\n\n2. **Exchange Request**:\n   - You have requested to exchange the following item:\n     - Hiking Boots, with the Item ID: 8118291112\n   - The replacement item you have selected is:\n     - Waterproof Hiking Boots, Size 12, Leather, with the Item ID: 8277474082\n   - Regarding the price difference, the new pair of boots is priced at $236.57, whereas the original pair was $260.56. As a result, you will receive a refund of $23.99, which will also be credited to your PayPal account.\n\nAt this point, I kindly ask that you review the information above to ensure that all details are accurate and reflect your intentions. If everything is correct, please let me know, and I will be happy to proceed with processing your return and exchange requests promptly. If you notice any discrepancies or if you have any additional questions, please do not hesitate to inform me so I can assist you further. Thank you for your attention, and I look forward to your confirmation."
},
\end{promptbox}

\end{document}